\documentclass[11pt]{article}

\usepackage{acl}

\usepackage{times}
\usepackage{latexsym}
\usepackage[T1]{fontenc}
\usepackage[utf8]{inputenc}
\usepackage{microtype}
\usepackage{inconsolata}
\usepackage{graphicx}
\usepackage{booktabs}
\usepackage{multirow}
\usepackage[table]{xcolor} 
\usepackage{tabularx}
\usepackage[most]{tcolorbox}
\usepackage{newunicodechar}
\usepackage[edges]{forest}
\usepackage{amsfonts}
\usepackage{array}
\usepackage{amsmath}
\usepackage{hyperref}
\usepackage{booktabs}
\usepackage{tabularx}
\usepackage{array}
\usepackage[table]{xcolor}

\newcommand{\paper}{\textit{Padārtha}}

\title{Padārtha: Ontology-Grounded Fine-Grained NER Benchmark for Classical Sanskrit}

\author{
  Sujoy Sarkar$^{1}$, 
  Pretam Ray$^{1}$, 
  Paramhans Shah$^{1}$, 
  Manoj Balaji Jagadeeshan$^{1}$, \\
  \textbf{Akash Gairola$^{2}$, 
  Arjuna S R$^{3}$, 
  Pawan Goyal$^{1}$} \\[0.5em]
  $^{1}$Indian Institute of Technology Kharagpur, India \\
  $^{2}$Central Sanskrit University, Devprayag, India \\
  $^{3}$Manipal Academy of Higher Education, Manipal, India
}

\begin{document}
\maketitle

\begin{abstract}
Annotation schemas are not neutral. When applied to classical literature, tag sets developed for modern journalistic texts impose source-culture definitions on texts they were never designed to describe. We instead ground a schema in the tradition of the text itself introducing \textit{Padārtha}, the first ontology-grounded fine-grained Named Entity Recognition (NER) benchmark for Sanskrit, built on the \textit{Mahābhārata} epic. Our tag set derives from \textit{Nyāya-Vaiśeṣika}, a classical Indian ontological system, yielding 18 fine-grained categories organized under 10 ontological nodes and mapped onto five standard coarse tags, ensuring interoperability with existing benchmarks. Expert annotators label over 12.6K entries from a scholarly index of named entities, linked to corresponding mentions in the \textit{Mahānāma} corpus, producing fine-grained annotations for 108,335 entity mentions across 73,632 verses, along with a 5,000-verse expert-verified test set sampled to stress rare mentions. We present the first systematic benchmarking of generative NER against traditional architectures for Sanskrit, finding that fine-tuned generative models perform comparably to task-specific systems. However, all systems show a sharp decline from coarse to fine granularity and struggle with out-of-entity mentions unseen during training. The limitation is not due to data scarcity alone, as fine-tuned models recall unseen entities far worse than seen ones and tend to default to the majority sense under lexical ambiguity.

\end{abstract}

\section{Introduction}
\label{sec:introduction}

As a core task in natural language understanding, Named Entity Recognition (NER) is widely adopted across domains such as news~\cite{tedeschi-navigli-2022-multinerd}, medical texts~\cite{biored-10.1093/bib/bbac282}, and social media~\cite{wnut-derczynski-etal-2017-results}. Despite its broad applicability, current systems perform well on coarse categories but struggle with nuanced or specialised entity types~\cite{fewnerd-ding-etal-2021-nerd,neretrieve-katz-etal-2023-neretrieve, multiconer-fetahu-etal-2023-multiconer}, with performance dropping further as categories move deeper into a hierarchy~\cite{neretrieve-katz-etal-2023-neretrieve}. These challenges have been studied almost exclusively in standard domains; literary text, where entities follow distinct distributions and narrative behaviour, remains comparatively neglected~\cite{literary-vala-etal-2015-mr,literary-bamman-etal-2019-annotated,silva-moro-2024-pportal,zhao-etal-2025-genwebnovel}.

Creating fine-grained NER datasets for literary texts, particularly classical corpora such as epics, religious works, and historical literature, is challenging because standard tagsets either lack sufficient granularity or are designed for modern, journalistic domains with limited domain-specific types~\cite{liu2021crossner}. These limitations are amplified in texts containing culturally specific items (CSIs). Schema like OntoNotes do not represent mythological entities~\cite{tedeschi-navigli-2022-multinerd}, while CoNLL-style tagsets remain overly coarse, relying on ill-defined catch-all categories such as MISC~\cite{mayhew-etal-2024-universal}. The difficulty is not that a label is missing but that the underlying partition of the world is wrong: ``Garu\d{d}a'', for instance, is simultaneously a bird, a deity, the son of Ka\'{s}yapa and Vinat\={a}, and the vehicle of Vi\d{s}\d{n}u, and HiNER~\cite{murthy-etal-2022-hiner} reports precisely this class of entity as resistant to assignment under \texttt{PERSON}. Consequently, if such schemes are not applied with due care, they risk overlooking the complexity of the target culture and imposing source-culture concepts onto it~\cite{hershcovich-etal-2022-challenges}.

Beyond schema limitations, annotated literary data are scarce~\cite{zhao-etal-2025-genwebnovel}, especially for historical texts where corpus construction requires costly expert knowledge~\cite{palladino2026more}. Unlike standardized news text, fiction exhibits substantial lexical variation with diverse, context-dependent naming conventions~\cite{han-etal-2021-fantasycoref}, and a few major entities typically dominate literary discourse~\cite{sundar-etal-2024-major}.  This skewed distribution leaves rare entities highly susceptible to out-of-entity (OOE) failures~\cite{jiang-etal-2025-mitigating}, where mention tokens at test time are entirely unseen during training.

This raises a question that the literary NER literature has not addressed: when a text's entities resist an imported tagset, is it better to extend that tagset ad hoc, or to ground the schema in a categorial system internal to the text's own intellectual tradition?
We argue for the latter. An \textit{emic} schema, one drawn from within the tradition, fixes category boundaries that were articulated independently of the annotation task, and so cannot be accused of being drawn to fit the data post hoc; this is the constructive counterpart to the warning of \citet{hershcovich-etal-2022-challenges}.

\begin{figure}[t]
\centering
\includegraphics[width=\columnwidth]{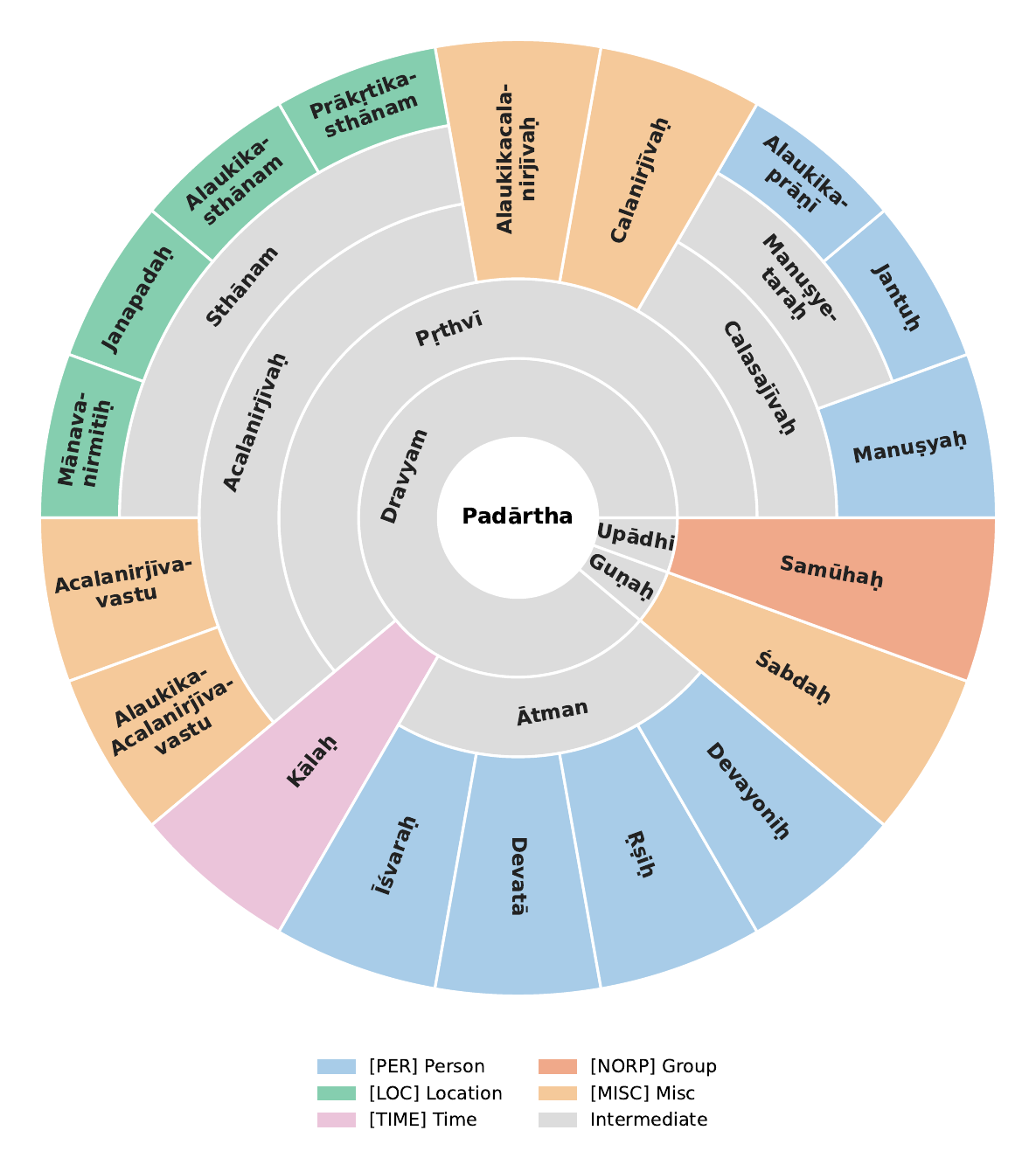}

\vspace{2pt}
\noindent\footnotesize\textit{padārtha} = ontological root; \textit{dravya} = substance; \textit{guṇa} = quality; \textit{upādhi} = incidental property; \textit{pṛthvī} = earth; \textit{ātman} = soul; \textit{sthānam} = places; \textit{calasajīvaḥ} = mobile living beings; \textit{acalanirjīvaḥ} = immobile non-living things; \textit{manuṣyetaraḥ} = non-human beings.

\caption{Nyāya-Vaiśeṣika ontology hierarchy underlying the tagset. Leaf nodes correspond to fine-grained categories (glosses and examples in Table~\ref{tab:tagsets_full}). Color indicates standard coarse category (legend).}
\label{fig:ontology_sunburst}
\end{figure}

To bridge these gaps and investigate the unique challenges of the literary domain, we introduce \paper, the first fine-grained NER benchmark in Sanskrit, based on the \textit{Mahābhārata}, one of the longest epics in world literature. Its vast narrative captures diverse naming conventions, while its composition in Sanskrit, a morphologically rich language, presents additional computational hurdles and establishes a typological contrast to standard English benchmarks. In this work, our core contributions are:


\paragraph{(1) Ontology-Grounded Hierarchical Tagset} We propose an annotation scheme grounded in Sanskrit linguistic extensions of Ny\={a}ya--Vai\'{s}e\d{s}ika (NV)~\cite{nair2013extended}, an ancient Indian ontological framework, to capture culturally specific items (CSIs) at the granularity the epic demands. The scheme has 18 fine-grained leaves under 10 ontology nodes, each mapping to one of five standard tags (\texttt{PER}, \texttt{LOC}, \texttt{NORP}, \texttt{MISC}, \texttt{TIME}), so it stays compatible with existing NER benchmarks (Figure~\ref{fig:ontology_sunburst}).

\paragraph{(2) Fine-Grained Sanskrit NER Dataset} We build \textit{Padārtha} based on the \textit{Mah\={a}n\={a}ma} corpus~\cite{sarkar-etal-2025-mahanama}, which pre-links mentions to S{\o}rensen's \textit{Index ~\cite{sorensen1904index}}. A Sanskrit scholar labeled 12.6K index entries from their descriptions using our schema; projecting these labels through the existing links gives 73,632 verses and 108,335 mentions. We then read and corrected a 5,000-verse test set by hand to serve as gold evaluation data.

\begin{table*}[t]
\centering
\small
\renewcommand{\arraystretch}{1.1}
\setlength{\tabcolsep}{5pt}
\definecolor{rowperson}{gray}{0.95}
\definecolor{rowlocation}{gray}{0.90}
\definecolor{rownorp}{gray}{0.90}
\definecolor{rowmisc}{gray}{0.95}
\definecolor{rowtime}{gray}{0.95}

\begin{tabularx}{\textwidth}{c l l p{3.5cm} l X}
\toprule
\textbf{SL} & \textbf{Coarse} & \textbf{Fine Tag} & \textbf{Gloss} & \textbf{Example} & \textbf{Sørensen Description} \\
\midrule

\rowcolor{rowperson}
1 & \textbf{Person} & \textit{īśvaraḥ} & The Trimūrti (Trinity) & Śiva & Great god (Mahādeva) \\
\rowcolor{rowperson}
2 & \textbf{Person} & \textit{devatā} & Gods & Indra & King of the gods \\
\rowcolor{rowperson}
3 & \textbf{Person} & \textit{ṛṣiḥ} & Sages & Vasiṣṭha & Celebrated celestial sage \\
\rowcolor{rowperson}
4 & \textbf{Person} & \textit{devayoniḥ} & Divine-origin beings & Citraratha & King of the Gandharvas \\
\rowcolor{rowperson}
5 & \textbf{Person} & \textit{manuṣyaḥ} & Humans & Arjuna & Third Pāṇḍava prince \\
\rowcolor{rowperson}
6 & \textbf{Person} & \textit{jantuḥ} & Animals & Sugrīva & One of Kṛṣṇa's horses \\
\rowcolor{rowperson}
7 & \textbf{Person} & \textit{alaukikaprāṇī} & Mythical creatures & Haṃsikā & A celestial cow \\

\midrule
\rowcolor{rowlocation}
8 & \textbf{Location} & \textit{prākṛtikasthānam} & Natural locations & Daṇḍakāraṇya & A forest \\
\rowcolor{rowlocation}
9 & \textbf{Location} & \textit{alaukikasthānam} & Mythical places & Gandharvaloka & World of the Gandharvas \\
\rowcolor{rowlocation}
10 & \textbf{Location} & \textit{janapadaḥ} & Geo-political entities & Aṅga & The Aṅga country \\
\rowcolor{rowlocation}
11 & \textbf{Location} & \textit{mānavanirmitaḥ} & Man-made structures & Agastyāśrama & Hermitage of Agastya \\

\midrule
\rowcolor{rowmisc}
12 & \textbf{Misc} & \textit{calanirjīvaḥ} & Mobile artifacts & Jaitra & A chariot \\
\rowcolor{rowmisc}
13 & \textbf{Misc} & \textit{acalanirjīvavastu} & Immobile artifacts & Asampṛṣṭha & An immobile entity/artifact \\
\rowcolor{rowmisc}
14 & \textbf{Misc} & \textit{alaukikacalanirjīvaḥ} & Mythical mobile artifacts & Puṣpaka & A celestial car \\
\rowcolor{rowmisc}
15 & \textbf{Misc} & \textit{alaukikacalanirjīvavastu} & Mythical immobile artifacts & Sudarśana & The discus of Kṛṣṇa \\
\rowcolor{rowmisc}
16 & \textbf{Misc} & \textit{śabdaḥ} & Texts and mantras & Nītiśāstra & Science of ethics \\

\midrule
\rowcolor{rownorp}
17 & \textbf{NORP} & \textit{samūhaḥ} & Nationalities, Religious, Groups & Ābhīra & People west of the Indus \\

\midrule
\rowcolor{rowtime}
18 & \textbf{Time} & \textit{kālaḥ} & Units of time & Bhādrapada & Name of a month \\

\bottomrule
\end{tabularx}

\caption{Complete coarse and mapped fine-grained Nyāya-Vaiśeṣika tagset, featuring examples and their canonical descriptions from the book \textit{Index to the Names in the Mahābhārata}~\cite{sorensen1904index}.}
\label{tab:tagsets_full}
\end{table*}

\paragraph{(3) Comprehensive Benchmarking} Motivated by the shift toward modern LLMs, we benchmark decoder-only LLMs under both parameter-efficient fine-tuning and few-shot in-context learning, contextualized against a CRF baseline, a fine-tuned encoder, and a generative encoder-decoder. Performance degrades consistently as categorical granularity increases, evaluated on a test split constructed to capture rare and unseen OOE mentions.

\paragraph{(4) Empirical and Theoretical Analysis of Fine-Grained Challenges}
We analyze the degradation at finer granularity along four dimensions: memorization vs.\ generalization, lexical ambiguity, information-theoretic difficulty, and qualitative error patterns. Fine-tuned models show substantially lower recall on unseen entities and default to the majority sense under ambiguity. Information-theoretic analysis confirms that context, rather than the entity itself, serves as the primary signal.

\begin{table*}[t]
\centering
\footnotesize
\setlength{\tabcolsep}{5pt}
\renewcommand{\arraystretch}{1.15}

\begin{tabularx}{\textwidth}{
>{\raggedright\arraybackslash}p{2.4cm}
>{\raggedright\arraybackslash}p{2.8cm}
>{\raggedright\arraybackslash}X}
\toprule
\rowcolor{gray!15}
\textbf{Format} &
\textbf{Used By} &
\textbf{Example Output (Mahābhārata CE Volume 1 Chapter 1 Verse 2)} \\
\midrule

\textbf{BIO Tags}
&
CRF, MuRIL, ByT5
&
loma (\textcolor{blue}{B-PER})
harṣaṇa (\textcolor{blue}{I-PER})
putra (\textcolor{blue}{I-PER}),
ugraśravāḥ (\textcolor{blue}{B-PER}),
sautiḥ (\textcolor{blue}{B-PER}),
paurāṇikā (\textcolor{blue}{B-PER}),
u (\textcolor{red}{O}),
naimiṣa (\textcolor{teal}{B-LOC}),
araṇye (\textcolor{red}{O}),
śaunakasya (\textcolor{blue}{B-PER})
\ldots
\\

\midrule

\textbf{Inline Bracketed}
&
Qwen2.5-7B,

Gemma-4-12B

(fine-tuning)
&
[
loma harṣaṇa putra
$\mid$
\textcolor{blue}{PER}
]
[
ugraśravāḥ
$\mid$
\textcolor{blue}{PER}
]
[
sautiḥ
$\mid$
\textcolor{blue}{PER}
]
[
paurāṇikā
$\mid$
\textcolor{blue}{PER}
]
u
[
naimiṣa
$\mid$
\textcolor{teal}{LOC}
]
araṇye
[
śaunakasya
$\mid$
\textcolor{blue}{PER}
]
\ldots
\\

\midrule

\textbf{Two-Stage Pipeline}
&
Qwen2.5-7B,

Gemma-4-12B,

Gemini 3 Flash

(few-shot)
&
\textbf{1. Mention Detection:}

{\ttfamily\small
@@loma harṣaṇa putra\#\#
@@ugraśravāḥ\#\#
@@sautiḥ\#\#
@@paurāṇikā\#\#
u
@@naimiṣa\#\#
araṇye
@@śaunakasya\#\#
\ldots
}

\vspace{2mm}

\textbf{2. Classification:}

{\ttfamily\small
[
\textcolor{blue}{"PER"},
\textcolor{blue}{"PER"},
\textcolor{blue}{"PER"},
\textcolor{blue}{"PER"},
\textcolor{teal}{"LOC"},
\textcolor{blue}{"PER"},
\ldots
]
}
\\

\bottomrule
\end{tabularx}

\caption{Example verse and output representation formats by model and training paradigm, shown on the same example verse (``\ldots sage Lomaharṣaṇa's son Ugraśravas [=Sauti], well-read in the Purāṇas, [came to] the Naimiṣa forest, [during] Śaunaka's sacrifice\ldots''). All formats are converted to a unified BIO representation for scoring. In the Two-Stage Pipeline row, \texttt{@@}\ldots\texttt{\#\#} mark candidate mentions from the detection stage, following the cascade design of \cite{luo-etal-2025-dynamicner}.}
\label{tab:ner_formats_iast}
\end{table*}

\section{Tagset Design and Dataset Construction}

\subsection{The Nyāya-Vaiśeṣika Ontology and NER Adaptation}

The Nyāya Vaiśeṣika (NV) school unifies Nyāya, founded by Gautama (c.\ 300 BCE) and centered on epistemology and logic, with Vaiśeṣika, founded by Kaṇāda, which develops a systematic ontology of existence~\cite{nair2013extended}. It classifies reality into seven \textit{padārthas}, of which \textit{dravya} (substance) is most relevant to Named Entity Recognition. The Vaiśeṣika Sūtra defines \textit{dravya} as \textit{kriyāguṇavat samavāyikāraṇam}, that which possesses actions and attributes and serves as an inherent cause (VS I.1.15)~\cite{sacredbooks_vol6_1911}, supporting the treatment of  entities as discrete objects with properties and relations.

Our fine-grained tagset draws directly on the extended Vaiśeṣika \textit{dravya} classification of \citet{nair2013extended}, which refines the classical scheme and adds \textit{alaukika} (celestial) nodes to accommodate entities in the Amarakośa lexicon. Fifteen of our eighteen categories (Table~\ref{tab:tagsets_full}) are inherited from this \textit{dravya} hierarchy, preserving category names and scope; the selection process is described in \S\ref{sec:tag_selection}. To cover entities outside the substance branch, we include \textit{samūhaḥ} and \textit{śabdaḥ} from the \textit{upādhi} (adventitious property) and \textit{guṇa} (quality) nodes, respectively. We further extend the framework with \textit{janapadaḥ} to expand the classification of places (\textit{sthānam}) beyond man-made and natural categories.

Unlike standard flat NER schemas, the NV-derived taxonomy is hierarchical, with each terminal label reached through intermediate categories rather than assigned independently (Figure~\ref{fig:ontology_sunburst}). For instance, \textit{manuṣyaḥ} (humans) follows substance $\rightarrow$ earth $\rightarrow$ mobile living beings $\rightarrow$ \textit{manuṣyaḥ}, while \textit{prākṛtikasthānam} (natural locations) follows substance $\rightarrow$ earth $\rightarrow$ immobile non-living things $\rightarrow$ places $\rightarrow$ \textit{prākṛtikasthānam}, sharing a common root before diverging. All eighteen terminal categories map to five coarse NER tags (Table~\ref{tab:tagsets_full}), preserving compatibility while retaining hierarchical structure.

\subsection{Base Resource: The Mahānāma Corpus}
\label{sec:base_resource}

We build \paper upon the \textit{Mahānāma} corpus~\cite{sarkar-etal-2025-mahanama}, the first large-scale Sanskrit entity linking dataset derived from the Indian epic \textit{Mahābhārata}. The epic's extended narrative poses significant entity resolution challenges due to the high variability and ambiguity of classical names~\cite{sarkar-etal-2025-mahanama}. \textit{Mahānāma} links entity mentions to canonical entities in a knowledge base, where each entity aliases is associated with a description from Sørensen’s \textit{Index to the Names in the Mahābhārata}~\cite{sorensen1904index}\footnote{\url{https://www.sanskrit-lexicon.uni-koeln.de/scans/INMScan/2020/web/index.php}}. These descriptions, covering approximately 12.6K entries, serve as the primary basis for our ontological annotation (Table~\ref{tab:tagsets_full}). \textit{Mahānāma} provides three coarse-grained tags (PER, LOC, and MISC); however, it does not provide any guidelines on classification and does not conduct any experiments on NER models.

\subsection{Tag Set}
\label{sec:tag_selection}

The extended NV ontology~\citet{nair2013extended} contains substantially more categories. We narrowed it to eighteen fine-grained tags (Table~\ref{tab:tagsets_full}) using two criteria. First, we retained only categories functioning as named entities rather than common nouns, excluding branches such as plants, minerals, the \textit{tejaḥ} branch, and all \textit{guṇaḥ} subdivisions except \textit{śabdaḥ}. Second, we collapsed distinctions with insufficient corpus support into their parent node,\footnote{E.g., the six habitat-based subdivisions of \textit{jantuḥ} and the \textit{apauruṣeyam}/\textit{pauruṣeyam} distinction within \textit{śabdaḥ}; the latter mirrors OntoNotes' single \texttt{WORK\_OF\_ART} tag, which similarly does not subdivide by origin~\cite{hovy-etal-2006-ontonotes}.} except where collapsing would merge semantically distinct siblings, in which case we retained the sparse node (e.g., \textit{mānavanirmitaḥ}, \textit{calanirjīvaḥ}, \textit{alaukikacalanirjīvaḥ}).

Our coarse tagset maps to standard NER schemas. We adopt \texttt{PERSON}, \texttt{LOCATION}, and \texttt{MISC} from the CoNLL convention~\cite{tjong-kim-sang-de-meulder-2003-introduction}, but omit \texttt{ORG}, since its standard definition, institutional or organizational entities, does not meaningfully apply to classical Sanskrit texts. We additionally introduce two coarse categories from OntoNotes~\cite{hovy-etal-2006-ontonotes}, \texttt{NORP} and \texttt{TIME}, to accommodate fine-grained NV categories (\textit{samūhaḥ}, \textit{kālaḥ}) that CoNLL's scheme would otherwise conflate with unrelated entities under tags such as \texttt{MISC} (\S\ref{sec:base_resource}). The resulting five-tag coarse scheme remains a strict refinement of PER/LOC/MISC and can be trivially collapsed back to it for compatibility with existing tools.

\subsection{Annotation}
Annotation was performed at the level of Index entries linked to Mahānāma mention spans. A doctoral researcher in Sanskrit assigned one of the eighteen NV derived fine grained tags to each of the 12.6K Index entries based on its description (Table~\ref{tab:tagsets_full}), consulting the original verse when needed. The assigned label was then inherited by all linked mentions. To account for potential omissions in Mahānāma’s mention marking~\cite{sarkar-etal-2025-mahanama}, the annotator additionally manually reviewed and corrected a 5,000 verse test set.

Unlike modern texts, classical Sanskrit narrative frequently identifies entities through epithets, derived forms, and relational compounds rather than repeating proper names. Since this is a common way of expressing entity mentions in classical texts, the Index treats such expressions as name references, and we followed the convention: \textit{derived forms} (e.g., \textit{Brāhmī} from Brahmā, \textit{Aindrī} from Indra), \textit{relational identifiers} (e.g., \textit{Kuntīputra}, "son of Kuntī"), and \textit{epithets} (e.g., \textit{Vṛkodara}-"he of the voracious appetite"-for Bhīma) are all classified under the entity they identify.

\subsubsection{Annotation Reliability}
\label{sec:iaa}

\begin{table}[h]
\centering
\small
\begin{tabular}{lccc}
\toprule
\textbf{Level} & \textbf{Agreement (\%)} & \textbf{$\kappa$} \\
\midrule
Coarse label & 91.03 & 0.728 \\
Fine label & 79.71 & 0.746 \\
\midrule
Span detection & \multicolumn{2}{c}{F1 = 0.811} \\
\bottomrule
\end{tabular}
\caption{Annotation reliability between the primary annotator and a second reviewer, over 1000 doubly annotated test verses.}
\label{tab:iaa}
\end{table}

We assess annotation reliability by comparing the primary annotator's labels against a postdoctoral Sanskrit grammarian with no prior exposure to the Index of Names, over 1,000 doubly annotated test verses (Table~\ref{tab:iaa}). Both granularities show substantial agreement \citep{landis1977measurement}. Coarse $\kappa$ is lower than fine $\kappa$ despite higher raw agreement, a known effect of label-distribution skew on chance-corrected measures.

\subsection{Dataset Statistics}
\label{sec:dataset_stats}

\begin{figure}[h]
\centering
\includegraphics[width=\columnwidth]{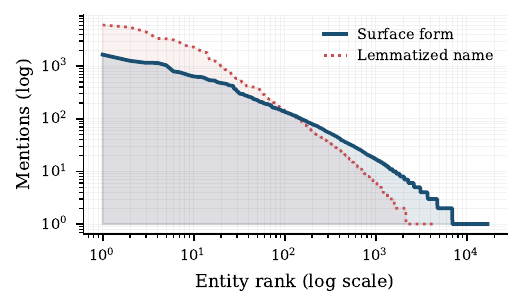}
\caption{Rank-frequency distribution of entity mentions}
\label{fig:name_freq_skew}
\end{figure}

The final dataset contains 108,335 entity mentions across eighteen fine-grained categories (Table~\ref{tab:fine_distribution}) and five coarse groupings (Table~\ref{tab:coarse_distribution}). Fgure~\ref{fig:name_freq_skew} shows the rank–frequency distribution of entity mentions, where both surface forms and their lemmatized counterparts exhibit a long-tailed skew, with surface forms spanning a wider rank range due to morphological variation.

\begin{table}[h]
\centering
\small
\setlength{\tabcolsep}{5pt}
\begin{tabular}{lr@{\hspace{12pt}}lr}
\toprule
\textbf{Category} & \textbf{Count} & \textbf{Category} & \textbf{Count} \\
\midrule
PERSON  & 88,930 & MISC     & 4,537 \\
NORP    & 9,416  & LOCATION & 4,798 \\
TIME    & 654    &          &       \\
\midrule
\textbf{Total} & \multicolumn{3}{r}{108,335 mentions} \\
\bottomrule
\end{tabular}
\caption{Distribution of coarse-grained entity categories.}
\label{tab:coarse_distribution}
\end{table}

\begin{table}[t]
\centering
\footnotesize
\renewcommand{\arraystretch}{1.08}
\setlength{\tabcolsep}{3pt}
\resizebox{\columnwidth}{!}{
\begin{tabular}{l r @{\hspace{10pt}} l r}
\toprule
\textbf{Tag} & \textbf{Count} & \textbf{Tag} & \textbf{Count} \\
\midrule
manuṣyaḥ & 51,969 & devatā & 11,149 \\
īśvaraḥ & 11,271 & samūhaḥ & 9,416 \\
ṛṣiḥ & 6,424 & devayoniḥ & 6,022 \\
prākṛtikasthānam & 3,284 & śabdaḥ & 3,132 \\
alaukikaprāṇī & 1,939 & alaukika\_acalanirjīvavastu & 856 \\
alaukikasthānam & 806 & janapadaḥ & 690 \\
kālaḥ & 654 & acalanirjīvavastu & 512 \\
jantuḥ & 156 & alaukikacalanirjīvaḥ & 20 \\
mānavanirmitaḥ & 18 & calanirjīvaḥ & 17 \\
\midrule
\textbf{Total} & \multicolumn{3}{r}{108,335 mentions} \\
\bottomrule
\end{tabular}
}
\caption{Distribution of fine-grained entity categories.}
\label{tab:fine_distribution}
\end{table}

\subsection{Data Split}
From 73,632 verses we selected 5,000-verse test set by sampling verses with rare names (appearing in $\leq$5 verses), ensuring stratified coverage across fine-grained types. Because selection is verse-level, non-rare entities also appear; overall, 73.14\% of unique surface forms, accounting for 50.47\% of all test mentions, are unseen during training. The train–test distribution across coarse- and fine-grained schemas is shown in Table~\ref{tab:train_test_distribution}.

\section{Experimental Setup}
\subsection{Models Used}
For Sanskrit NER, we consider models spanning four architectural paradigms: a non-neural CRF~\cite{sutton2012introduction-crf} baseline trained on surface features; MuRIL-large, a multilingual encoder pretrained on Indic languages~\cite{khanuja2021muril}; ByT5-Sanskrit, an encoder-decoder pretrained on Sanskrit segmentation and morphology~\cite{nehrdich-etal-2024-one}; Qwen2.5-7B-Instruct~\cite{qwen2025qwen25technicalreport} and Gemma-4-12B~\cite{gemmateam2024gemmaopenmodelsbased}, decoder-only models evaluated under both fine-tuning and few-shot prompting; and Gemini 3 Flash, a closed-source frontier model~\cite{geminiteam2024gemini15unlockingmultimodal} evaluated few-shot only (full settings in Appendix~\ref{app:hyperparams}).

\paragraph{Baseline.} We train a linear-chain CRF on shallow surface features (the token itself, character prefixes/suffixes, and a $\pm1$ token context window), establishing a non-neural baseline.

\paragraph{Encoder Fine-tuning.} MuRIL-large is trained using the SpanMarkerNER framework\footnote{\url{https://github.com/tomaarsen/SpanMarkerNER}}, following standard span-based BIO sequence labeling (learning rate 1e-5, batch size 8, up to 10 epochs).

\paragraph{Encoder-Decoder Fine-tuning.} We use ByT5-Sanskrit, pretrained on Sanskrit segmentation and morphological tagging. To test whether prior training on related Sanskrit tasks transfers to NER, we keep the model's original \texttt{token\_tag} generation format, substituting NER labels (learning rate 3e-4, batch size 8, 15 epochs).

\paragraph{Decoder Fine-tuning.} Qwen2.5-7B-Instruct and Gemma-4-12B-it are fine-tuned via LoRA (full hyperparameters in Appendix~\ref{app:hyperparams}) using the Inline Bracketed output format (Table~\ref{tab:ner_formats_iast}; full templates in Appendix~\ref{app:prompts}, Figure~\ref{fig:prompts_fine_tuning}), which \citet{zhan2026assessment} found to outperform alternative output formats for LLM-based NER.

\paragraph{In-Context Learning.} Following CascadeNER's two-stage strategy~\cite{luo-etal-2025-dynamicner}, we adopt a mention-detection-then-classification pipeline for Qwen2.5-7B-Instruct, Gemma-4-12B-it, and Gemini 3 Flash. Stage 1 marks candidate spans inline within the full sentence; Stage 2 classifies each span as a JSON label list, with 5-shot demonstrations selected via $k$-nearest-neighbor retrieval using ByT5-Sanskrit encoder embeddings (full prompts in Appendix~\ref{app:prompts}, Figure~\ref{fig:prompts_few_shot}).

\paragraph{Script.} ByT5-Sanskrit uses IAST transliteration\footnote{\url{https://en.wikipedia.org/wiki/International_Alphabet_of_Sanskrit_Transliteration}}, consistent with its original pretraining; all other models use Devanagari.

\paragraph{Prompt Design}
Fine-tuning uses a single instruction specifying the full tagset (gloss and examples per tag), span-tagging rules, and strict output formatting.Few-shot instead uses the two-stage cascade described above, with an added \textsc{Unknown} label in Stage 2 to reject spans Stage 1 over-generates. Both paradigms share the same entity-span rules (proper names, derived names, relational identifiers, epithets). Full prompts in Appendix~\ref{app:prompts}.

\subsection{Evaluation Metrics}

All model outputs are converted to a unified BIO token-tag representation and evaluated using \texttt{seqeval}\footnote{\url{https://pypi.org/project/seqeval/}} with the IOB2 scheme in strict (CoNLL-style) mode, requiring exact span boundary and label match. We report Precision, Recall, and F1, at both coarse and fine granularity. Given severe class imbalance (\S\ref{sec:dataset_stats}), we report both Micro-F1, dominated by frequent categories, and Macro-F1, which weights all categories equally and better reflects performance on sparse tags.

\section{Results}
\label{sec:results}
\begin{table*}[t]
    \centering
    \small 
    \setlength{\tabcolsep}{4pt}
    \begin{tabular}{llll ccc ccc}
    \toprule
    \multirow{2}{*}{\textbf{Methodology}} & \multirow{2}{*}{\textbf{Architecture}} & \multirow{2}{*}{\textbf{Model}} & \multirow{2}{*}{\textbf{Output Format}} & \multicolumn{3}{c}{\textbf{Micro Average (\%)}} & \multicolumn{3}{c}{\textbf{Macro Average (\%)}} \\
    \cmidrule(lr){5-7} \cmidrule(lr){8-10}
    & & & & \textbf{P} & \textbf{R} & \textbf{F1} & \textbf{P} & \textbf{R} & \textbf{F1} \\
    \midrule
    
    \multicolumn{10}{c}{\textbf{Coarse-Grained}} \\
    \midrule
    Baseline & -- & CRF & BIO & 78.83 & 40.05 & 53.12 & 74.46 & 32.08 & 44.37 \\
    \midrule
    
    \multirow{4}{*}{Fine-tuning} 
     & Encoder & MuRIL & BIO & \textbf{79.84} & 52.09 & 63.04 & \textbf{75.46} & 48.77 & \textbf{58.56} \\
     & Enc-Dec & ByT5 & BIO & 78.96 & 51.85 & 62.60 & 72.53 & 44.96 & 54.63 \\
     & Decoder & Qwen-2.5-7B & \parbox{2cm}{In. Bracketed} & 79.38 & 53.37 & 63.83 & 58.75 & 41.75 & 48.63 \\
     & Decoder & Gemma-4-12B & \parbox{2cm}{In. Bracketed} & 78.69 & 51.91 & 62.55 & 73.29 & 41.89 & 53.10 \\
    \midrule
    
    \multirow{3}{*}{Few-Shot} 
     & Decoder & Qwen-2.5-7B & Two-Stage & 17.85 & 24.07 & 20.50 & 17.27 & 18.83 & 15.12 \\
     & Decoder & Gemma-4-12B & Two-Stage & 48.01 & 42.81 & 45.26 & 31.33 & 40.28 & 32.48 \\
     & Closed-Source & Gemini 3 Flash & Two-Stage & 66.34 & \textbf{76.70} & \textbf{71.14} & 47.66 & \textbf{69.09} & 55.05 \\
    
    \midrule
    \midrule
    
    \multicolumn{10}{c}{\textbf{Fine-Grained}} \\
    \midrule
    Baseline & -- & CRF & BIO & 69.73 & 33.87 & 45.59 & 58.37 & 23.93 & 32.80 \\
    \midrule
    
    \multirow{5}{*}{Fine-tuning} 
     & Encoder & MuRIL & BIO & \textbf{70.16} & 45.23 & 55.00 & 59.38 & 34.78 & \textbf{41.82} \\
     & Enc-Dec & ByT5 & BIO & 63.21 & 42.37 & 50.73 & 45.07 & 29.30 & 34.37 \\
     & Decoder & Qwen-2.5-7B & \parbox{2cm}{In. Bracketed} & 67.05 & 47.96 & 55.92 & 52.24 & 36.53 & 41.77 \\
     & Decoder & Gemma-4-12B & \parbox{2cm}{In. Bracketed} & 69.89 & 43.38 & 53.53 & 49.61 & 28.07 & 34.23 \\
    \midrule
    
    \multirow{3}{*}{Few-Shot} 
     & Decoder & Qwen-2.5-7B & Two-Stage & 25.38 & 8.53 & 12.77 & 13.24 & 4.81 & 6.41 \\
     & Decoder & Gemma-4-12B & Two-Stage & 38.04 & 31.19 & 34.28 & 22.68 & 24.00 & 20.50 \\
     & Closed-Source & Gemini-3 Flash & Two-Stage & 58.35 & \textbf{65.94} & \textbf{61.91} & 36.52 & 48.47 & 39.80 \\
    \bottomrule
    \end{tabular}
    \caption{Precision, Recall, and F1 for all models at coarse-grained and fine-grained granularity, grouped by training paradigm (fine-tuning vs.\ few-shot) and architecture. Bold marks the best score in each column within a granularity.}
    \label{tab:comp_full_metrics_stacked}
\end{table*}

\begin{table}[t]
    \centering
    \small
    \begin{tabular}{lccc}
    \toprule
    \textbf{Model} & \textbf{P} & \textbf{R} & \textbf{F1} \\
    \midrule
    CRF & 85.86 & 43.63 & 57.86 \\
    MuRIL & 86.45 & 56.40 & 68.27 \\
    ByT5 & 85.99 & 56.48 & 68.18 \\
    Gemini-3 Flash & 71.14 & 82.24 & 76.29 \\
    Qwen-2.5-7B (FT) & 82.62 & 55.55 & 66.43 \\
    Gemma-4-12B (FT) & 85.11 & 56.14 & 67.66 \\
    Qwen-2.5-7B (Fewshot) & 36.61 & 49.38 & 42.05 \\
    Gemma-4-12B (Fewshot) & 58.74 & 52.37 & 55.37 \\
    \bottomrule
    \end{tabular}
    \caption{Mention Detection performance across models (coarse-grained setup). (\S\ref{sec:ablation}).}
    \label{tab:mention_detection_summary}
\end{table}

\subsection{Overall Performance}
Fine-tuned neural models substantially outperform the CRF baseline (Table~\ref{tab:comp_full_metrics_stacked}), whereas few-shot decoder-only models (Qwen-FS, Gemma-FS) underperform it at both granularities. Fine-tuning consistently improves over few-shot for the same backbone. Compared to MuRIL, gains are mixed: MuRIL retains the highest Macro-F1, while only Qwen-FT narrowly leads on Micro-F1 (per-label breakdown in Appendix~\ref{app:per-label-heatmaps}).

Mention detection performance (Table~\ref{tab:mention_detection_summary}) confirms that the primary recall bottleneck across models stems fundamentally from span extraction failures driven by the Out of Entity challenge (\S\ref{sec:introduction}, \S\ref{sec:seen_unseen}).  Gemini 3 Flash is the exception, with recall exceeding precision. All models degrade from coarse to fine prediction, more in Macro F1 (avg.\ $-13.8$pp) than Micro F1 (avg.\ $-9.0$pp), reflecting low support categories and increased disambiguation demands.

Gemini 3 Flash achieves the highest Micro F1 at both granularities, surpassing all fine tuned models, while MuRIL retains the best Macro F1. Given the Mahābhārata’s public availability, some pretraining exposure is possible, though our ontology differs from existing annotations.

\subsection{Seen vs.\ Unseen Entity Recall}
\label{sec:seen_unseen}
We define seen and unseen based on exact surface-form overlap between training and test data. All models recall seen entities far better than unseen ones, confirming the Out-of-Entity challenge (full results in Table~\ref{tab:seen_unseen}). CRF and fine-tuned models show the largest seen/unseen disparities, indicating heavy reliance on memorization, whereas Gemini retains substantially higher unseen recall (55.6--67.2\%).

\begin{table}[h]
\centering
\small
\setlength{\tabcolsep}{5pt}
\begin{tabular}{lrrrr}
\toprule
& \multicolumn{2}{c}{Coarse (Recall \%)} & \multicolumn{2}{c}{Fine (Recall \%)} \\
\cmidrule(lr){2-3} \cmidrule(lr){4-5}
Model & Seen & Unseen & Seen & Unseen \\
\midrule
CRF            & 71.9 & 8.8  & 63.7 & 4.6  \\
MuRIL          & 80.6 & 24.0 & 72.5 & 18.3 \\
ByT5           & 81.9 & 22.4 & 71.3 & 13.7 \\
Qwen-FT        & 81.8 & 25.1 & 75.2 & 21.1 \\
Gemma-FT       & 77.0 & 27.2 & 68.4 & 18.8 \\
Qwen-FS        & 30.4 & 17.8 & 11.9 & 5.2  \\
Gemma-FS       & 59.3 & 26.5 & 46.5 & 16.1 \\
Gemini-3~Flash & 86.3 & 67.2 & 76.4 & 55.6 \\
\bottomrule
\end{tabular}
\caption{Recall (\%) for seen vs.\ unseen entities, based on exact surface-form overlap with training data.}
\label{tab:seen_unseen}
\end{table}

\subsection{Impact of Lexical Ambiguity}
\textit{Mahānāma} paper~\cite{sarkar-etal-2025-mahanama} identifies contextual ambiguity as a central challenge, where identical expressions refer to different entities depending on context. To assess its impact on NER, we analyze performance on minority-sense homonyms, where the correct fine-grained tag contradicts the most frequent surface-form label (Appendix~\ref{app:homonymy}). We found 1227 minority sense mentions in test data. Fine-tuned models (MuRIL, Gemma-FT, Qwen-FT) consistently fall into a majority-sense trap, with wrong-label rates (27.4\%--32.6\%) matching or exceeding correct predictions. 

\subsection{V-information Analysis}
\label{sec:vinfo}

Following \citet{ma-etal-2023-towards}, we train MuRIL-based entity-only and context-only classifiers to compute pointwise $\mathcal{V}$-information for mentions in the test set. To prevent extreme outlier bias, we exclude highly sparse classes when calculating the dataset-level mean $\mathcal{V}$-information (Table~\ref{tab:vinfo}). Evaluating these metrics reveals that entity $\mathcal{V}$-information is negative while context $\mathcal{V}$-information is positive at both granularities, establishing context as the primary signal.


\section{Error Analysis}
\subsection{Quantitative Analysis}
\begin{figure}[t]
\centering
\includegraphics[width=\columnwidth]{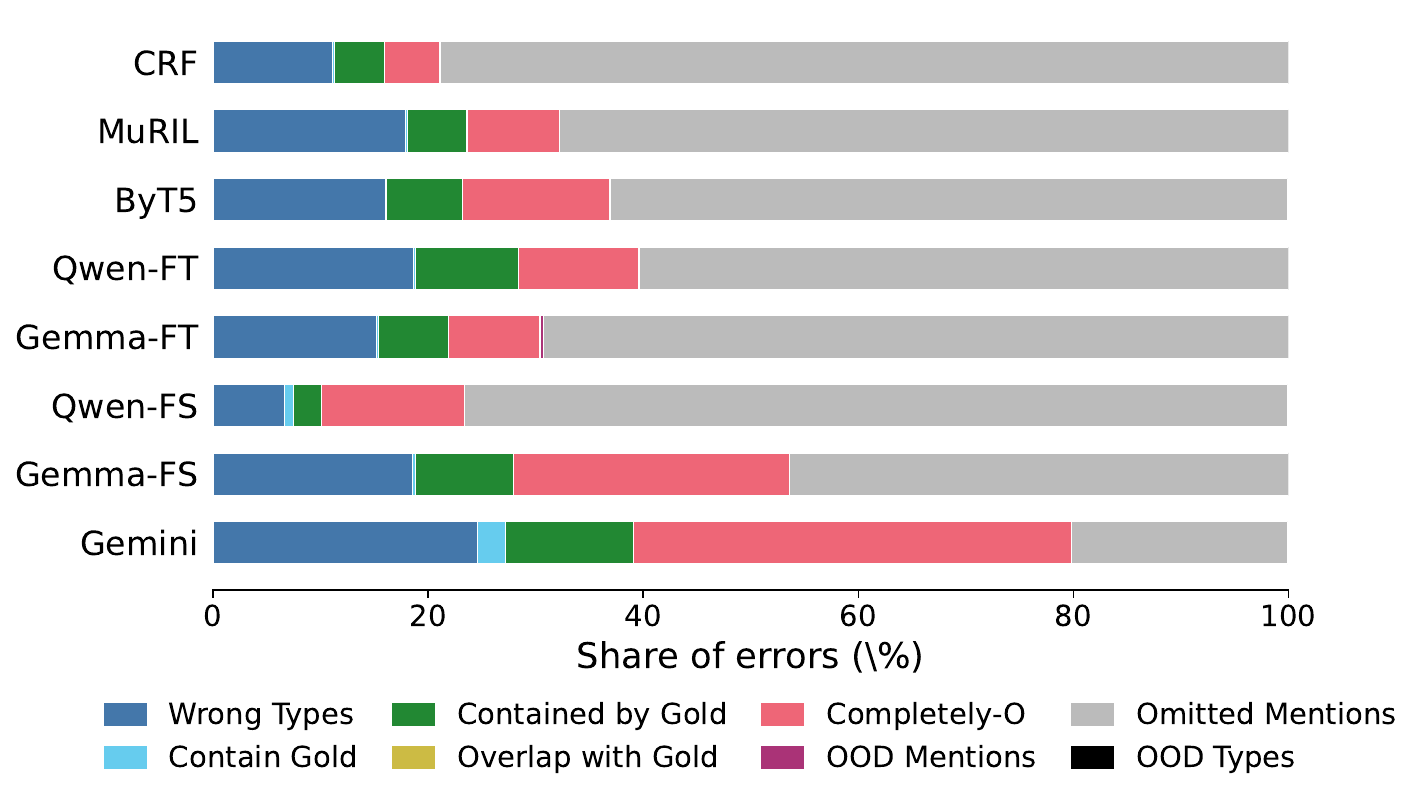}
\caption{Fine-grained error type distribution by model}
\label{fig:error_taxonomy_fine}
\end{figure}

We classify predictions into eight error categories following the taxonomy by \cite{xie-etal-2023-empirical} (see definition in Appendix~\ref{app:error_taxonomy}), with full distributions reported in Table~\ref{tab:error_distribution_full} and summarized (fine-grained) in Figure~\ref{fig:error_taxonomy_fine}.

Error profiles split cleanly by training paradigm at the coarse-grained level: gradient-trained models (fine-tuned and CRF) are dominated by Omitted Mentions, while few-shot models show comparatively more Completely-O errors. This pattern holds at both granularities (Table~\ref{tab:error_distribution_full}) and across architectures, suggesting the effect stems from the training paradigm itself.

Wrong-Type errors rise consistently from coarse to fine across nearly every model, showing that fine-grained classification is inherently harder than coarse. Qwen-FS is the sole exception, with Wrong-Type errors dropping from 16.8\% to 6.7\% possibly because its Omitted Mentions rate jumps sharply (26.5\% to 76.3\%). We found it driven by its heavy reliance on the \texttt{UNKNOWN} fallback tag we used in few-shot classification step (75.1\% of fine-grained predictions, vs.\ 8.0\% for Gemma-FS and 2.3\% for Gemini). Boundary errors are similarly asymmetric, with under-extension dominating over over-extension across all models, an effect most pronounced in few-shot models.

\subsection{Manual Analysis}
\label{sec:qualitative_errors}
To understand the error patterns in Table~\ref{tab:error_distribution_full}, we manually reviewed a sample of model predictions.
Wrong Type errors tend to arise mostly in two situations. The first is confusion between fine-grained categories that share the same coarse type. For example, mythical mobile and immobile objects are often mixed up, so weapons get tagged as \textit{alaukikacalanirjīvaḥ} (mythical mobile artifact) instead of \textit{alaukika\_acalanirjīvavastu} (mythical immobile artifact). The second situation is category assignment driven by an entity's relationships rather than the entity itself. Gemini, for instance, tags the goddess Dānu as \textit{devayoniḥ} (semi-divine being) in one context and \textit{devatā} in another, depending on which relation, mother of the Dānavas or daughter of Dakṣa, is most salient in the surrounding text, rather than the entity's own consistent type. 
Under-extension errors tend to arise when a model stops reading at the first word boundary instead of continuing to the full entity span. For example, Gemini tags only \textit{baka} instead of the full phrase \textit{baka vadhaḥ}, or dropping \textit{-nandana} from \textit{keśava-nandana} which stems from frequenct use of compounds and relational identifiers as mentions in classical Sanskrit texts.

\section{Ablation Study}
\label{sec:ablation}

\begin{table}[t]
\centering
\small
\setlength{\tabcolsep}{4pt}
\renewcommand{\arraystretch}{0.95}
\begin{tabular}{@{}lcc@{}}
\toprule
\textbf{Setting} & \textbf{Micro F1} & \textbf{Macro F1} \\
\midrule
Coarse, oracle detection    & 66.1 & 50.1 \\
Fine, w/ coarse (oracle)    & 47.5 & 34.9 \\
Fine, w/o coarse (oracle)   & 45.7 & 29.4 \\
\bottomrule
\end{tabular}
\caption{Oracle-detection ablations for Gemma.}
\label{tab:oracle-ablation}
\end{table}

\paragraph{Coarse-level oracle detection:} We ablate the few-shot Gemma setup by replacing predicted spans with gold spans and rerunning coarse classification to isolate detection from classification error. F1 rises from 45.26/32.48 to 66.1/50.1 (micro/macro), showing detection as the bottleneck. Classification with oracle detection exceeds fine-tuned Gemma's micro F1 (62.55) but trails its macro F1 (53.10 vs.\ 50.1). MISC shows high recall but low precision (73.4/13.0), acting as a default under uncertainty.

\paragraph{Fine-level oracle detection:} We repeat the oracle-detection ablation at the fine-grained level using gold spans, comparing two settings: predicting the fine label given the gold coarse label (w/ coarse) versus predicting it from the gold span alone (w/o coarse). Both (47.5/34.9 and 45.7/29.4 F1; Table~\ref{tab:oracle-ablation}) trail coarse-oracle despite identical oracle detection, showing granularity, not only detection, drives the difficulty. Sparse classes like \textit{calanirjīvaḥ} and \textit{mānavanirmitaḥ} collapse toward zero F1, likely from too few instances for effective KNN retrieval.


\section{Conclusion}
We built \paper:Ontology-Grounded Fine-Grained NER Benchmark for Classical Sanskrit, by grounding the tagset in Ny\={a}ya--Vai\'{s}e\d{s}ika rather than adapting a schema designed for modern news. The resulting 18 categories sit under 10 hierarchical nodes and collapse to five standard coarse tags, so the data remains usable with existing tools. Labeling 12.6K entries of S{\o}rensen's \textit{Index} and projecting these through the \textit{Mah\={a}n\={a}ma} links gives 108,335 mentions over 73,632 verses, with a 5,000-verse test set corrected by hand.

Our benchmarking reveals that while task-specific designs on Sanskrit are now comparable with refined generative models, neither model performs well in the fine-grained context. The $\mathcal{V}$-information analysis points to why: fine categories carry more of their signal in the surrounding context than in the mention itself, which is exactly what a memorizing model cannot exploit. 

\section*{Limitations}
All experiments use machine-segmented text; we do not address the additional challenges posed by sandhi in Sanskrit, which we leave for future work. Our benchmark is built on a classical epic in verse form, so applicability to prose texts may be limited and would require further investigation, for instance using poetry-to-prose conversion techniques. We do not address the class imbalance present in the dataset, which we leave for future investigation. The source corpus also contains some OCR and automatic segmentation errors; assessing their impact is out of scope for this work, which instead focuses on a standardized classification guideline grounded in the ontology.

\section*{Ethics Statement}

The annotations in this work are derived from published, copyright-free sources and a publicly available corpus~\cite{sarkar-etal-2025-mahanama, sorensen1904index}. All resources utilized have been appropriately cited. The dataset, including annotations, is constructed entirely from existing literary sources; no explicit bias analysis has been performed. The dataset, annotations, and code will be released under a CC-0 license.
Annotation was carried out by a doctoral researcher in Sanskrit, who is also an author of this paper, and who assigned fine-grained labels to each of the 12.6K Index entries and manually reviewed and corrected the 5,000-verse test set. Annotation reliability was assessed by comparing the primary annotator's labels against a postdoctoral Sanskrit grammarian with no prior exposure to the Index of Names, over 1,000 doubly annotated test verses; this review was conducted as a courtesy and was not compensated. The dataset does not contain any personal or sensitive information. 

\section*{Acknowledgments}
This work was supported in part by the GCP Research Grant for Gemma and the National Language Translation Mission (NLTM): Bhashini project of the Government of India. We also thank Dr. Soorya A. P., Postdoctoral Fellow at Manipal Academy of Higher Education, Manipal, India, for independently reviewing the annotations to assess annotation reliability.

\section*{AI Assistance}

AI assistants were used during the writing process to refine textual clarity, grammar, and phrasing. AI assistants were also used during development to assist with debugging code used in the experimental pipeline.

\bibliography{custom}

\appendix

\section{Dataset Statistics}
\label{app:dataset_stats}
Table~\ref{tab:train_test_distribution} reports the number of entity mentions per category in the training and test splits, for both coarse- and fine-grained schemas.
\begin{table}[h]
    \centering
    \small
    \begin{tabular}{lrr}
    \toprule
    \textbf{Entity Class} & \textbf{Train} & \textbf{Test} \\
    \midrule
    \multicolumn{3}{c}{\textbf{Coarse-Grained}} \\
    \midrule
    person & 72,503 & 16,427 \\
    norp & 7,319 & 2,097 \\
    misc & 3,577 & 960 \\
    location & 2,879 & 1,919 \\
    time & 558 & 96 \\
    \midrule
    \textbf{Total (Coarse)} & \textbf{86,836} & \textbf{21,499} \\
    \midrule
    \midrule
    \multicolumn{3}{c}{\textbf{Fine-Grained}} \\
    \midrule
    manuṣyaḥ & 45,377 & 6,592 \\
    devatā & 8,217 & 2,932 \\
    samūhaḥ & 7,319 & 2,097 \\
    īśvaraḥ & 7,228 & 4,043 \\
    ṛṣiḥ & 5,195 & 1,229 \\
    devayoniḥ & 5,080 & 942 \\
    śabdaḥ & 2,498 & 634 \\
    prākṛtikasthānam & 1,721 & 1,563 \\
    alaukikaprāṇī & 1,305 & 634 \\
    alaukika\_acalanirjīvavastu & 649 & 207 \\
    alaukikasthānam & 628 & 178 \\
    kālaḥ & 558 & 96 \\
    janapadaḥ & 523 & 167 \\
    acalanirjīvavastu & 408 & 104 \\
    jantuḥ & 101 & 55 \\
    alaukikacalanirjīvaḥ & 15 & 5 \\
    calanirjīvaḥ & 7 & 10 \\
    mānavanirmitaḥ & 7 & 11 \\
    \midrule
    \textbf{Total (Fine)} & \textbf{86,836} & \textbf{21,499} \\
    \bottomrule
    \end{tabular}
    \caption{Distribution of entity mentions across the training and test sets for both coarse-grained and fine-grained labeling schemas.}
    \label{tab:train_test_distribution}
\end{table}

\section{Implementation Details}
\label{app:hyperparams}
\paragraph{CRF.} We train a linear-chain CRF \texttt{sklearn-crfsuite}) using L-BFGS optimization $c_1=0.1$, $c_2=0.1$, max 100 iterations, all possible transitions enabled), on shallow surface features: token identity, character prefixes/suffixes (1--3 characters), a $\pm1$ token context window, and sentence-boundary markers.

\paragraph{MuRIL.} We fine-tune \texttt{google/muril-large-cased} using the SpanMarkerNER framework (max sequence length 512, entity max length 8), with learning rate 1e-5, warmup ratio 0.1, batch size 8, up to 10 epochs with early stopping (patience 3), fp16 precision.

\paragraph{ByT5.} We fine-tune \texttt{chronbmm/sanskrit5-multitask} for sequence-to-sequence tag generation, using learning rate 3e-4, batch size 8, 15 epochs, max sequence length 512, linear scheduler with 100 warmup steps.

\paragraph{Qwen2.5-7B Fine-Tuning.} We fine-tune \texttt{Qwen/Qwen2.5-7B-Instruct} via LoRA (rank 32, $\alpha=64$) using the LLaMA-Factory framework\footnote{\url{https://github.com/hiyouga/LlamaFactory}}, learning rate 2e-5, effective batch size 16, 3 epochs, bf16 precision, max sequence length 1536, cosine scheduler with warmup ratio 0.05.

\paragraph{Gemma-4-12B Fine-Tuning.} We fine-tune \texttt{google/gemma-4-12b-it} via LoRA (rank 32, $\alpha=64$) using Unsloth\footnote{\url{https://unsloth.ai/}} (chosen as unavailability of model in LLaMA-Factory at the time of experiment and the mode for memory efficiency at this model size), learning rate 1e-4, effective batch size 16, 3 epochs, cosine scheduler with warmup ratio 0.05, bf16 precision, max sequence length 1536, AdamW optimizer, weight decay 0.01.

\paragraph{Qwen2.5-7B and Gemma-4-12B Few-Shot.} bf16 precision, temperature 1.0, top-$p$ 0.95.

\paragraph{Computational Budget.} All fine-tuning and inference for MuRIL, ByT5, Qwen2.5-7B, and Gemma-4-12B were performed on a single machine with 2$\times$ NVIDIA L40 GPUs (46GB each).

\section{Prompt Templates}
\label{app:prompts}
Figure~\ref{fig:prompts_fine_tuning} shows the full instruction templates used for supervised fine-tuning, at both fine-grained and coarse-grained levels. Figure~\ref{fig:prompts_few_shot} shows the two-stage few-shot cascade prompts. Few-shot inference uses the same tagset definitions within a two-stage cascade: a mention-extraction prompt (favoring recall) followed by a classification prompt assigning each detected span a category.

\begin{figure*}[t]
\centering
\small

\begin{tcolorbox}[colback=blue!5, colframe=black, title=Fine-Grained Fine-Tuning Instruction, fonttitle=\bfseries]
\textbf{You are an expert Sanskrit linguist performing Named Entity Recognition (NER) on the Mahābhārata.} Identify ALL named entities in the Sanskrit sentence. Rewrite the sentence with inline tags using the exact format: \texttt{[Entity Text | fine\_tag]}.

\textbf{Entity Types:} Use ONLY the 18 tags below.

\textit{Person / Beings:} (1) \textit{manuṣyaḥ} -- mortal humans (Arjuna, Duryodhana); (2) \textit{jantuḥ} -- ordinary named animals (Sugrīva, Aśvatthāman); (3) \textit{alaukikaprāṇī} -- mythical/divine creatures (Ananta, Garuḍa); (4) \textit{īśvaraḥ} -- Trimūrti and avatars (Brahmā, Viṣṇu, Śiva, Kṛṣṇa); (5) \textit{devatā} -- standard gods (Indra, Bhūmi, Vāyu); (6) \textit{ṛṣiḥ} -- sages/ascetics (Vasiṣṭha, Viśvāmitra); (7) \textit{devayoniḥ} -- semi-divine/demons (Kubera, Ghaṭotkaca).

\textit{Locations:} (8) \textit{prākṛtikasthānam} -- natural places (Daṇḍakāraṇya, Gaṅgā); (9) \textit{alaukikasthānam} -- mythical worlds/heavens/hells (Gandharvaloka, Kālasāhvaya); (10) \textit{janapadaḥ} -- geo-political areas (Aṅga, Avantī, Hastināpura); (11) \textit{mānavanirmitaḥ} -- man-made structures (Lākṣāgṛham, Agastyāśrama).

\textit{Artifacts \& Texts:} (12) \textit{calanirjīvaḥ} -- mundane mobile artifacts (Jaitra); (13) \textit{alaukikacalanirjīvaḥ} -- mythical mobile artifacts (Puṣpaka); (14) \textit{acalanirjīvavastu} -- mundane stationary objects (Asampṛṣṭha); (15) \textit{alaukika\_acalanirjīvavastu} -- mythical/celestial stationary objects (Bhārgavāstra, Sudarśana); (16) \textit{śabdaḥ} -- named texts/mantras (Yajus, Rāmāyaṇa).

\textit{Time:} (17) \textit{kālaḥ} -- specific units of time/eras (Bhādrapada, Ardhamāsa).

\textit{Groups:} (18) \textit{samūhaḥ} -- specific named groups/clans (Ābhīra, Amarāḥ).

\textbf{Indirect Entity Spans.} Standard proper nouns are always tagged; additionally tag the entire word/phrase when an entity is identified indirectly: (1) \textit{Derived Names} -- grammatically created from a name (Brāhmī); (2) \textit{Relational Identifiers} -- identifies via relationship (Kuntīputra, Kāśipati); (3) \textit{Epithets} -- descriptive titles used as names (Vṛkodara); 
\textbf{Critical Rules.} (1) Format strictly as \texttt{[Entity | tag]}, no spaces around the pipe. (2) Tag specifics only -- never generic nouns, unless acting as a proper-name substitute for a specific character (e.g., \textit{ācārya} for Droṇa). (3) No nested tags -- if a compound acts as a single identifier, tag the entire compound once. (4) Maintain exact  spelling, sandhi, and segmentation outside brackets.
\end{tcolorbox}

\vspace{4mm}

\begin{tcolorbox}[colback=green!5, colframe=black, title=Coarse-Grained Fine-Tuning Instruction, fonttitle=\bfseries]
\textbf{You are an expert Sanskrit linguist performing Named Entity Recognition (NER) on the Mahābhārata.} Identify ALL named entities in the Sanskrit sentence. Rewrite the sentence with inline tags using the exact format: \texttt{[Entity Text | LABEL]}.

\textbf{Entity Types:} Use ONLY the 5 broad tags below. (1) \textbf{PER} (Person/Being) -- all sentient beings, including gods, sages, humans, specific animals, and mythical creatures (Brahmā, Kṛṣṇa, Arjuna, Vasiṣṭha, Sugrīva, Ananta); (2) \textbf{LOC} (Location) -- all places, including natural features, mythical realms, kingdoms, and man-made structures (Gaṅgā, Gandharvaloka, Hastināpura, Lākṣāgṛham); (3) \textbf{MISC} (Miscellaneous) -- inanimate artifacts, vehicles, divine weapons, and texts (Puṣpaka, Bhārgavāstra, Sudarśana, Rāmāyaṇa); (4) \textbf{TIME} -- specific periods, eras, or units of time (Kṛtayuga, Bhādrapada); (5) \textbf{NORP} (Groups) -- specific named collectives, lineages, tribes, or religious groups (Ābhīra, Amarāḥ).

\textbf{Indirect Entity Spans.} Standard proper nouns are always tagged; additionally tag the entire word/phrase when an entity is identified indirectly: (1) \textit{Derived Names} -- grammatically created from a name (Brāhmī); (2) \textit{Relational Identifiers} -- identifies via relationship (Kuntīputra, Kāśipati); (3) \textit{Epithets} -- descriptive titles used as names (Vṛkodara);

\textbf{Critical Rules.} (1) Format strictly as \texttt{[Entity | tag]}, no spaces around the pipe. (2) Tag specifics only -- never generic nouns, unless acting as a proper-name substitute for a specific character (e.g., \textit{ācārya} for Droṇa). (3) No nested tags -- if a compound acts as a single identifier, tag the entire compound once. (4) Maintain exact  spelling, sandhi, and segmentation outside brackets.
\end{tcolorbox}

\caption{Fine-tuning instruction templates for fine-grained (top) and coarse-grained (bottom) Sanskrit NER.}

\label{fig:prompts_fine_tuning}
\end{figure*}

\begin{figure*}[t]
\centering
\small

\begin{tcolorbox}[colback=blue!5, colframe=black, title=Stage 1: Mention Extraction Prompt (Few-Shot), fonttitle=\bfseries]
\textbf{You are an expert Sanskrit linguist performing Named Entity Recognition (NER) on the Mahābhārata.}

\textbf{Task Requirements:} (1) Surround entity spans using \texttt{@@} and \texttt{\#\#}. (2) Surround the output sentence with \texttt{\$\$} and \texttt{\$\$}. (3) Do NOT change, remove, or reorder any tokens. (4) Identify ALL possible named entities. Do NOT miss any entity. (5) When in doubt, mark it as an entity. Prefer higher recall over precision.

\textbf{Entity Types:} (1) \textbf{PER} (Person/Being) -- all sentient beings, including gods, sages, humans, specific animals, and mythical creatures (Brahmā, Kṛṣṇa, Arjuna, Vasiṣṭha, Sugrīva, Ananta); (2) \textbf{LOC} (Location) -- all places, including natural features, mythical realms, kingdoms, and man-made structures (Gaṅgā, Gandharvaloka, Hastināpura, Lākṣāgṛham); (3) \textbf{MISC} (Miscellaneous) -- inanimate artifacts, vehicles, divine weapons, and texts (Puṣpaka, Bhārgavāstra, Sudarśana, Rāmāyaṇa); (4) \textbf{TIME} -- specific periods, eras, or units of time (Kṛtayuga, Bhādrapada); (5) \textbf{NORP} (Groups) -- specific named collectives, lineages, tribes, or religious groups (Ābhīra, Amarāḥ).

\textbf{Indirect Entity Spans.} Standard proper nouns are always tagged; additionally tag the entire word/phrase when an entity is identified indirectly: (1) \textit{Derived Names} -- grammatically created from a name (Brāhmī); (2) \textit{Relational Identifiers} -- identifies via relationship (Kuntīputra, Kāśipati); (3) \textit{Epithets} -- descriptive titles used as names (Vṛkodara); 

\textbf{Critical Rules.} (1) Tag specifics only -- never generic nouns, unless acting as a proper-name substitute for a specific character (e.g., \textit{ācārya} for Droṇa). (2) No nested tags -- if a compound acts as a single identifier, tag the entire compound once. (3) Maintain exact spelling, sandhi, and segmentation outside brackets.

\textbf{IMPORTANT:} Output ONLY the tagged sentence wrapped in \texttt{\$\$\dots\$\$}. Nothing else.
\end{tcolorbox}

\vspace{4mm}

\begin{tcolorbox}[colback=orange!5, colframe=black, title=Stage 2: Classification Prompt (Few-Shot), fonttitle=\bfseries]
\textbf{You are an expert Sanskrit linguist performing Named Entity Recognition (NER) on the Mahābhārata. Classify the detected entities into the following labels:}

\textbf{Entity Types:} (1) \textbf{PER} (Person/Being) -- all sentient beings, including gods, sages, humans, specific animals, and mythical creatures (Brahmā, Kṛṣṇa, Arjuna, Vasiṣṭha, Sugrīva, Ananta); (2) \textbf{LOC} (Location) -- all places, including natural features, mythical realms, kingdoms, and man-made structures (Gaṅgā, Gandharvaloka, Hastināpura, Lākṣāgṛham); (3) \textbf{MISC} (Miscellaneous) -- inanimate artifacts, vehicles, divine weapons, and texts (Puṣpaka, Bhārgavāstra, Sudarśana, Rāmāyaṇa); (4) \textbf{TIME} -- specific periods, eras, or units of time (Kṛtayuga, Bhādrapada); (5) \textbf{NORP} (Groups) -- specific named collectives, lineages, tribes, or religious groups (Ābhīra, Amarāḥ); (6) \textbf{UNKNOWN} -- use this label if the span is clearly not a named entity.

\textbf{Output Format:} Return ONLY a JSON list of \{entity\_type\}, using k-nearest-neighbor--retrieved examples as in-context demonstrations. No explanations or additional text.
\end{tcolorbox}

\caption{Two-stage few-shot pipeline prompts: mention extraction (Stage 1) followed by classification (Stage 2).}
\label{fig:prompts_few_shot}
\end{figure*}

\section{Error Taxonomy}
\label{app:error_taxonomy}
We categorize sequence tagging discrepancies into eight distinct error types to diagnose typological, boundary, and detection failures:
\begin{itemize}
\item \textbf{Wrong Type:} The predicted boundary is exact, but the assigned category is incorrect.
\item \textbf{OOD Type (Out-of-Domain):} The predicted category falls outside the predefined label schema.
\item \textbf{Contained by Gold (Under-extension):} The prediction only captures a sub-part of the true entity span.
\item \textbf{Contain Gold (Over-extension):} The prediction completely covers the true entity but incorrectly includes adjacent non-entity text.
\item \textbf{Overlap with Gold:} A partial match where neither the predicted nor the true span fully contains the other.
\item \textbf{Completely-O (False Positive):} The predicted span has zero overlap with any ground-truth entity.
\item \textbf{Omitted Mention (False Negative):} A ground-truth entity is entirely missed by the model.
\item \textbf{OOD Mention (Hallucinated):} The model predicts an entity using a text string that does not physically exist in the input verse.
\end{itemize}
Table~\ref{tab:error_distribution_full} reports the complete error distribution across all models at both coarse and fine granularity, underlying the summary in Figure~\ref{fig:error_taxonomy_fine}. 
\begin{table*}[t]
\centering
\small
\setlength{\tabcolsep}{4pt}
\begin{tabular}{llrrrrrrrr}
\toprule
\textbf{Model} & \textbf{Gran.} & \textbf{Wrong} & \textbf{Contain} & \textbf{Contained} & \textbf{Overlap} & \textbf{Completely-O} & \textbf{OOD} & \textbf{Omitted} & \textbf{OOD} \\
 & & \textbf{Types} & \textbf{Gold} & \textbf{by Gold} & \textbf{with Gold} & & \textbf{Ment.} & \textbf{Mentions} & \textbf{Types} \\
\midrule
CRF & Coarse & 5.6\% & 0.3\% & 5.1\% & 0.0\% & 5.9\% & 0.0\% & 83.1\% & 0.0\% \\
CRF & Fine & 11.1\% & 0.2\% & 4.6\% & 0.0\% & 5.2\% & 0.0\% & 78.9\% & 0.0\% \\
\midrule
MuRIL & Coarse & 8.2\% & 0.2\% & 6.6\% & 0.0\% & 9.8\% & 0.0\% & 75.2\% & 0.0\% \\
MuRIL & Fine & 17.9\% & 0.2\% & 5.5\% & 0.0\% & 8.6\% & 0.0\% & 67.8\% & 0.0\% \\
\midrule
ByT5 & Coarse & 8.6\% & 0.3\% & 6.7\% & 0.0\% & 10.2\% & 0.0\% & 74.2\% & 0.0\% \\
ByT5 & Fine & 16.0\% & 0.1\% & 7.1\% & 0.0\% & 13.7\% & 0.0\% & 63.0\% & 0.0\% \\
\midrule
Qwen-FT & Coarse & 3.8\% & 0.5\% & 5.8\% & 0.1\% & 14.0\% & 4.5\% & 71.3\% & 0.0\% \\
Qwen-FT & Fine & 19.0\% & 0.2\% & 9.7\% & 0.0\% & 11.3\% & 0.0\% & 59.8\% & 0.0\% \\
\midrule
Gemma-FT & Coarse & 7.8\% & 0.3\% & 7.4\% & 0.0\% & 10.4\% & 0.4\% & 73.5\% & 0.0\% \\
Gemma-FT & Fine & 15.4\% & 0.2\% & 6.5\% & 0.0\% & 8.2\% & 0.3\% & 69.5\% & 0.0\% \\
\midrule
Qwen-FS & Coarse & 16.8\% & 1.9\% & 5.0\% & 0.0\% & 49.6\% & 0.0\% & 26.5\% & 0.0\% \\
Qwen-FS & Fine & 6.7\% & 0.9\% & 2.6\% & 0.0\% & 13.5\% & 0.0\% & 76.3\% & 0.0\% \\
\midrule
Gemma-FS & Coarse & 11.4\% & 0.4\% & 10.9\% & 0.0\% & 32.2\% & 0.0\% & 45.1\% & 0.0\% \\
Gemma-FS & Fine & 18.8\% & 0.3\% & 9.2\% & 0.0\% & 26.1\% & 0.0\% & 45.5\% & 0.0\% \\
\midrule
Gemini & Coarse & 11.6\% & 3.4\% & 14.7\% & 0.1\% & 51.4\% & 0.0\% & 18.8\% & 0.0\% \\
Gemini & Fine & 24.9\% & 2.7\% & 12.0\% & 0.0\% & 41.4\% & 0.0\% & 19.0\% & 0.0\% \\
\bottomrule
\end{tabular}
\caption{Error distribution across models and granularities. Percentages are relative to total errors.}
\label{tab:error_distribution_full}
\end{table*}

\subsection{Per-Label Performance Heatmaps}
\label{app:per-label-heatmaps}
\begin{figure*}[t]
    \centering
    \includegraphics[width=0.95\textwidth]{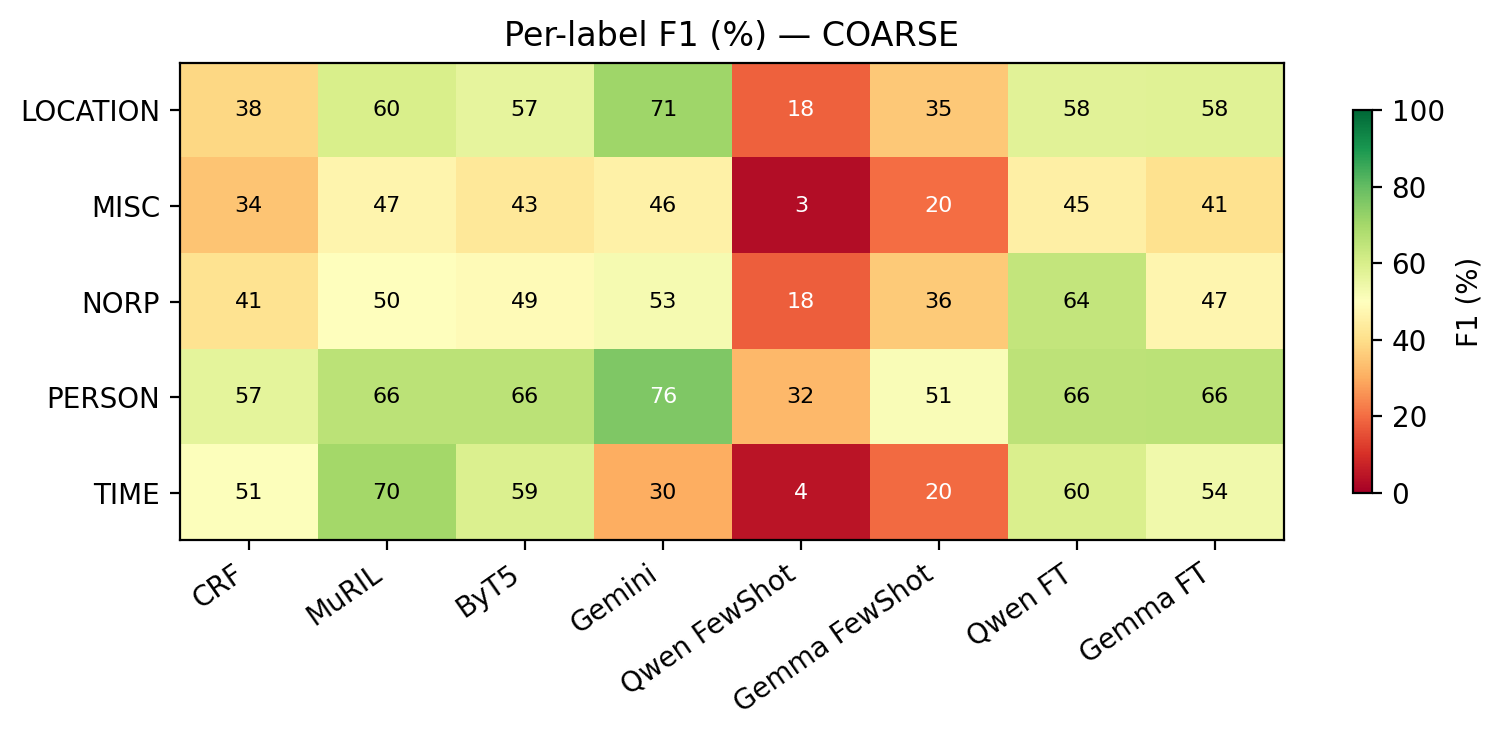}
    \\[6pt]
    \includegraphics[width=0.95\textwidth]{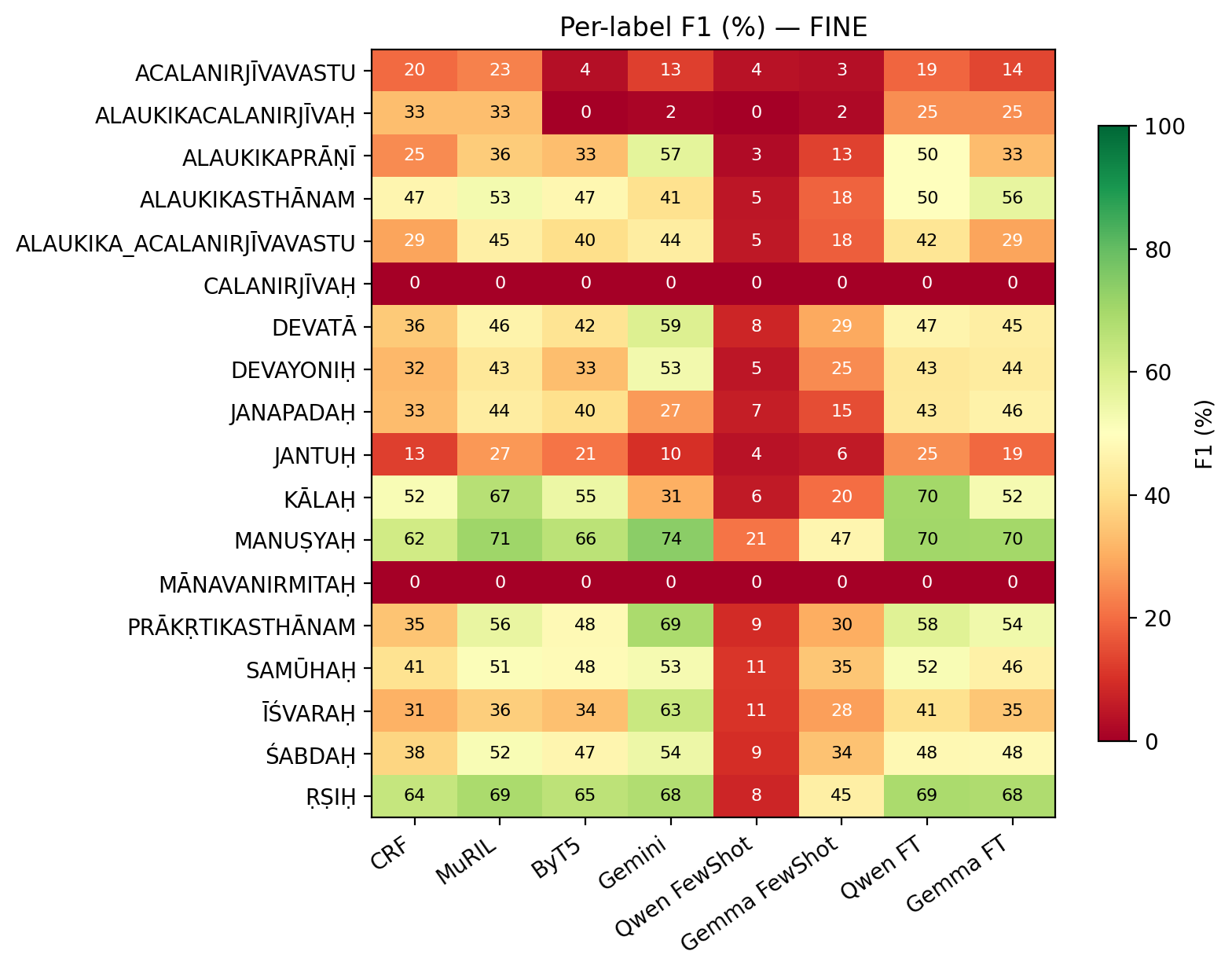}
    \caption{Per-label F1 (\%). Top: coarse-grained level. Bottom: fine-grained level (18 categories).}
    \label{fig:heatmap-combined}
\end{figure*}

To complement the aggregate micro/macro F1 scores reported in the main text, we provide a per-label breakdown of F1 performance across all eight evaluated models, at both the coarse-grained (5-category) and fine-grained (18-category) levels. Figure~\ref{fig:heatmap-combined} shows coarse-grained and fine-grained results; darker green indicates higher F1, darker red
indicates lower F1, with the exact value printed inside each cell. 



\section{V-Information}

We measure test dataset difficulty using $\mathcal{V}$-usable information, adapted for NER dataset bias analysis by \citet{ma-etal-2023-towards}. Following their method, we decouple entity and context by constructing two datasets from our NER data: an entity-only dataset (using only the entity span) and a context-only dataset (masking the entity with \texttt{[MASK]}, retaining the surrounding sentence). We fine-tune MuRIL-large separately on each to predict entity type.

For each test instance $(x,y)$, pointwise $\mathcal{V}$-information (PVI) is
\[
\text{PVI}(x \to y) = -\log_2 g[\emptyset](y) + \log_2 g'[x](y),
\]
where $g[\emptyset]$ is a null-input baseline estimated from training label frequencies, and $g'[x]$ is the entity-only or context-only model. Entity and Context $\mathcal{V}$-information are the mean PVI across test instances for each model.

Table~\ref{tab:vinfo} reports results at both granularities. Both show negative Entity $\mathcal{V}$-information and positive Context $\mathcal{V}$-information, indicating classification relies more on context than entity regularities. This gap is larger at fine granularity, suggesting fine-grained distinctions are more context-dependent.

\begin{table}[h]
\centering
\small
\begin{tabular}{lrr}
\toprule
\textbf{Granularity} & \textbf{Entity} & \textbf{Context} \\
                     & $\mathcal{V}$-Info & $\mathcal{V}$-Info \\
\midrule
Coarse (5 tags)        & $-0.18$ & $+0.41$ \\
Fine-grained (18 tags) & $-0.31$ & $+1.19$ \\
\bottomrule
\end{tabular}
\caption{$\mathcal{V}$-information for entity vs. context classifiers.}
\label{tab:vinfo}
\end{table}


\section{Minority-Sense Evaluation}
\label{app:homonymy}
\begin{table}[h]
\centering
\small
\begin{tabular}{lrrr}
\toprule
Model & Correct & Wrong & N.Det. \\
\midrule
CRF            & 8.3\%  & 25.7\% & 66.0\% \\
Gemini-3~Flash & 59.5\%  & 27.1\% & 13.3\% \\
Qwen-FT        & 31.8\%  & 32.1\% & 36.0\% \\
Qwen-FS        & 4.8\%  & 6.7\% & 88.5\% \\
MuRIL          & 24.2\%  & 32.6\% & 43.2\% \\
Gemma-FT       & 25.3\%  & 27.4\% & 47.4\% \\
Gemma-FS       & 17.7\%  & 28.9\% & 53.4\% \\
\bottomrule
\end{tabular}
\caption{Fine-grained accuracy on minority-sense homonyms. N.Det.\ = Not detected.}
\label{tab:minority_sense}
\end{table}

\section{Related Works}
The current state of Sanskrit Named Entity Recognition (NER) is marked by a transition toward large-scale data creation, yet a critical gap remains. Most existing resources are either coarse grained or classified as silver standard datasets derived through automated methods, and comprehensive benchmarking is largely absent. Early literary efforts, such as the pre-annotation based approach using the Śrīmad Bhāgavatam~\cite{sarkar-etal-2023-pre}, rely on a corpus of 18,000 verses and identify entities through a semi-automated heuristic that compares transliterated Sanskrit with English translations. This work does not employ traditional neural NER models, instead relying on string matching algorithms such as Jaro Winkler similarity to generate entity suggestions for annotators. The dataset is limited to coarse grained tags, specifically Person, Location, and Miscellaneous, and excludes organizational entities due to their absence in spiritual texts.

Similarly, the Mahānāma project~\cite{sarkar-etal-2025-mahanama}, derived from the Mahābhārata, identifies 109,000 mentions but focuses primarily on Entity Discovery and Linking and coreference resolution. Its experiments evaluate coreference and entity linking models rather than dedicated NER classification systems. It follows a similar coarse grained schema and lacks standardized annotation guidelines for Sanskrit, relying on an expert curated name index to define entities.

To address the scarcity of training data, recent frameworks have introduced large scale silver standard corpora, though these often lack the linguistic precision of manually annotated gold data. The Naamah corpus~\cite{p2026naamahlargescalesynthetic} provides a benchmark with 102,942 sentences and 127,397 total entities, including 90,452 Person, 22,290 Location, and 14,655 Organization tags. This dataset is used to benchmark transformer architectures, comparing XLM RoBERTa Base with the parameter efficient IndicBERTv2. However, it is synthetically generated using the Sarvam M model via DBpedia seeding and may inherit biases from prompting templates.

In the multilingual setting, SampurNER~\cite{kaushik2026sampurner} and TAFSIL~\cite{tafsil} extend fine grained NER to Sanskrit within broader Indian language initiatives, but both remain silver standard due to their methodologies. SampurNER employs the Entity Anchored Machine Translation framework to translate the English FewNERD dataset into Sanskrit, yielding 152,269 entities and evaluating models such as mBERT and IndicBERTv2. TAFSIL uses distant supervision by linking Wikipedia with Wikidata to construct datasets across multiple taxonomies, and evaluates noise aware models including LITE and DECENT with encoders such as MuRIL, XLM RoBERTa, and mDeBERTa.

Despite advances in scale, granularity, and model evaluation, a major gap remains in the availability of a domain native, gold standard Sanskrit dataset that does not rely on translation or synthetic generation, along with the absence of systematic and comparable benchmarking across NER approaches.

\end{document}